\documentclass[letterpaper, 10 pt, conference]{ieeeconf}  

\IEEEoverridecommandlockouts                              

\usepackage{amsmath} 
\usepackage{amssymb}  

\usepackage[pdftex]{graphicx}       
\usepackage[dvipsnames]{xcolor}
\usepackage{algorithm}
\usepackage[noend]{algpseudocode}
\usepackage{float}
\usepackage{siunitx}
\usepackage{hyperref}
\makeatletter
\let\NAT@parse\undefined
\makeatother
\usepackage[numbers]{natbib}

\newcommand{\eref}[1]{(\ref{#1})}
\newcommand{\secref}[1]{Section~\ref{#1}}
\newcommand{\tabref}[1]{Table~\ref{#1}}

\newcommand{\figref}[1]{Fig.~\ref{#1}}

\newcommand{\myparagraph}[1]{\vspace{0.03in}\noindent\textbf{#1}}
\newcommand*{\Cdot}{\raisebox{-0.25ex}{\scalebox{1.75}{$\cdot$}}}

\newboolean{draft-mode}
\setboolean{draft-mode}{true}
\newcommand{\sidenote}[1]{\ifthenelse{\boolean{draft-mode}}{\marginpar{\tiny\raggedright\textsf{\hspace{0pt}#1}}}{}}
\DeclareRobustCommand{\arnote}[1]{\ifthenelse{\boolean{draft-mode}}{\textcolor{blue}{\textbf{AR: #1}}}{}}
\DeclareRobustCommand{\ncdnote}[1]{\ifthenelse{\boolean{draft-mode}}{\textcolor{green}{\textbf{NCD: #1}}}{}}

\title{\LARGE \bf Regrasping by Fixtureless Fixturing\vspace{-3mm}}
\author{\authorblockN{Nikhil Chavan-Dafle and Alberto Rodriguez}
\authorblockA{Massachusetts Institute of Technology\\
{\tt\small \{nikhilcd, albertor\}@mit.edu}\vspace{10mm}}
\includegraphics[scale=1]{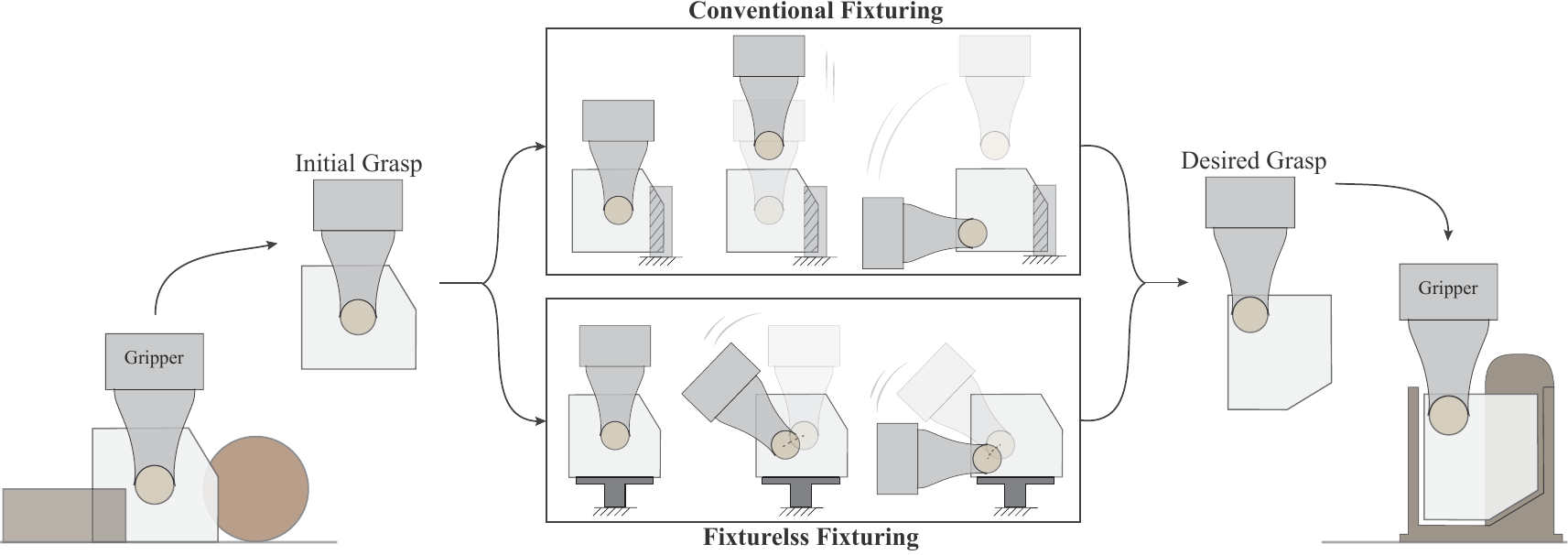}
}

\begin{document}

\maketitle
\thispagestyle{empty}
\pagestyle{empty}

\begin{abstract}
This paper presents a fixturing strategy for regrasping that does not require a physical fixture. To regrasp an object in a gripper, a robot pushes the object against external contact/s in the environment such that the external contact keeps the object stationary while the fingers slide over the object. We call this manipulation technique \emph{fixtureless fixturing}.
Exploiting the mechanics of pushing, we characterize a convex polyhedral set of pushes that results in fixtureless fixturing. These pushes are robust against uncertainty in the object inertia, grasping force, and the friction at the contacts. We propose a sampling-based planner that uses the sets of robust pushes to rapidly build a tree of reachable grasps. A path in this tree is a pushing strategy, possibly involving pushes from different sides, to regrasp the object. 
We demonstrate the experimental validity and robustness of the proposed manipulation technique with different regrasp examples on a manipulation platform. Such fast and flexible regrasp planner facilitates versatile and flexible automation solutions.  
\end{abstract}

\section{Introduction}
\label{sec:intro}

The need for regrasp is ubiquitous in assembly operations. In-hand manipulation skills allow one to pick up a part, reorient it in the hand, and then use/insert/assemble it. Robots, especially those with parallel-jaw grippers, lack the dexterity to autonomously adjust the grasp on an object. They are often assisted by systems such as part feeders and fixtures. Part feeders take care of reorienting the parts, while customized fixtures allow robots to regrasp through place-and-pick maneuvers. This approach leads to highly-customized, non-flexible, and expensive automation solutions.

This paper presents an in-hand manipulation technique that allows a robot with a simple parallel-jaw gripper to regrasp an object without any customized hardware. The robot pushes the object against external contact/s in the environment such that the object sticks to the external contact while the fingers slide over the object. Since the object is held stationary in the environment without any fixtures, we call this approach \emph{fixtureless fixturing}.

Fixtureless fixturing is a variant of prehensile pushing that is robust against uncertainties in the system parameters. In our prior work on prehensile pushing, we present an idea of manipulating an object with external pushes~\cite{ChavanDafle2014,ChavanDafle2015a,ChavanDafle2017,ChavanDafle2018a}. However, there we assume precise knowledge of system parameters such as the object inertia, grasping force, and friction coefficients at the fingers and at the external contacts. In practice, there is always some uncertainty in the knowledge of these parameters. Based on the mechanics of pushing, we define in this paper a subspace of robust prehensile pushes and refer to it as a \emph{robust motion cone}. We show that the robust motion cone is invariant to all the system parameters listed above except friction at the external contact/s.

\begin{figure*}
\centering
 \includegraphics[scale=0.98]{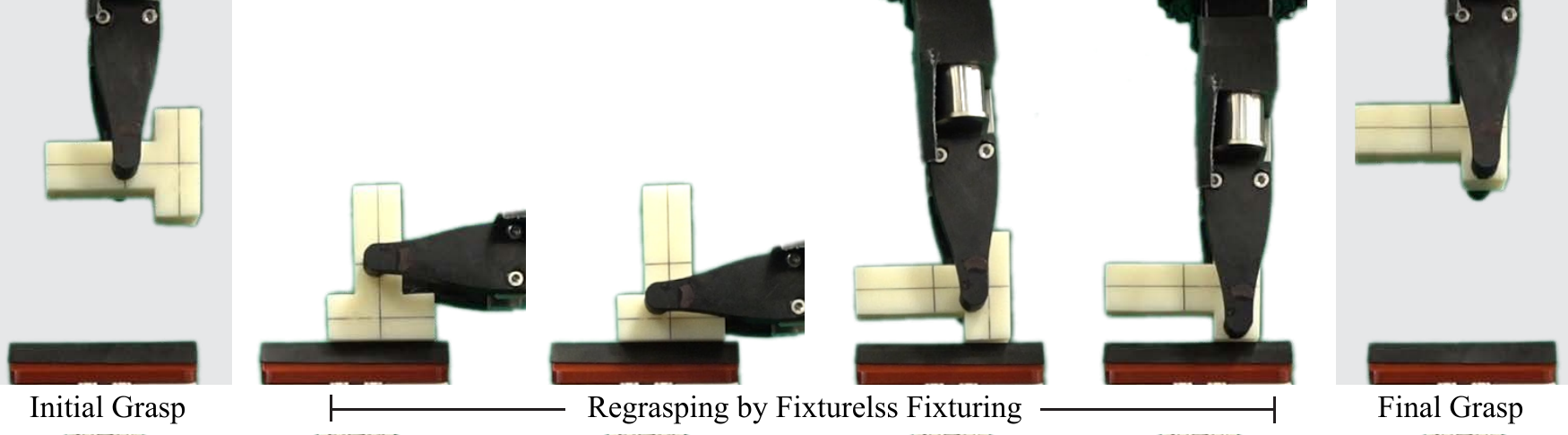}
\caption{An example of regrasping using fixtureless fixturing. A T-shaped object is regrasped in a gripper by pushing it against an edge contact in the environment. Note that during the regrasp process, the object sticks to the environment while the fingers slide on it.}
    \label{fig:tpush_exp}
\end{figure*} 

We present a sampling-based planning framework that uses robust motion cones to build long pushing strategies for regrasping. Exploiting the knowledge of the motion cones for dynamics propagation, the planner rapidly builds a tree of reachable grasps. A path is this tree is a pushing strategy, possibly involving pushes from different sides, to manipulate the object in the grasp. These pushing strategies force the object to the desired pose in the grasp irrespective of any uncertainty in the object inertia, grasping force, friction at the fingers. \figref{fig:tpush_exp} shows one such pushing strategy to manipulate a T-shaped object.

For experimental validation, we consider different examples of regrasping performed on a manipulation platform well-instrumented to capture the pose of the object and the robot. We show that the pushing strategies generated by our planner consistently achieve the desired regrasp and are robust against the uncertainties in the system parameters.

The efficient computation of the motion cones and their application to propagate the dynamics allows our planner to generate pushing strategies in less than a second. Moreover, the planner can use and reuse a simple feature in the environment to manipulate many different objects. Such a fast and robust framework for regrasping by fixtureless fixturing takes us a step closer towards a versatile and flexible method for in-hand manipulation for industrial automation. 

\section{Related Work}
\label{sec:related}
Fixturing is one of the widely used practices in industry for applications such as machining, assembly, and inspection. Researchers have extensively studied the principles and designs for fixturing~\cite{american1962handbook, Asada85, Chou89}. However, the role of friction in fixturing is often neglected or considered secondary to that of the kinematics.

One of the very few works that describe the use of force analysis with friction in the context of fixture planning is by Hong and Cutkosky~\cite{Hong91}. 
They point out that fixture planning, similar to grasp planning, is primarily considered from the kinematic perspective only. 
They advocate the analysis of the mechanics of fixturing and use the limit surface model~\cite{GoyalPhD89} to capture the force-motion relationship at the contacts between the object and the fixture.
Their work focuses on the stability of a fixtured part against the forces exerted by a tool and provides estimates for the clamping force required to prevent the part from sliding. Using similar contact mechanics tools, the work presented in this paper studies the controlled sliding of the object in the grasp. With the fixtureless fixturing, we demonstrate the  application of the mechanics of frictional contact for fixturing.


The work presented in this paper is motivated by our recent work on stable prehensile pushing~\cite{ChavanDafle2018a}. In that work, we present a planner to regrasp an object in a gripper using the external pushes for which the object sticks to the pusher. To find feasible stable prehensile pushes, the planner evaluates a dynamics check for multiple random push directions. This dynamics check depends on the knowledge of the object inertia, grasping force, and coefficient of friction at the fingers and at the features in the environment. Consequently, the stable prehnsile pushing is sensitive to these parameters. In contrast, the fixtureless fixturing presented in this paper uses only the stable prehensile pushes whose dynamics is invariant to these parameters except for the friction at the features in the environment. 
Moreover, we demonstrate a method to compute bounds on the valid robust stable pushes in the form of robust motion cones. The knowledge of the motion cones avoids the need of dynamics check and speeds up the regrasp planning.

The idea of robust motion cones is inspired by the original concept of the motion cone presented by Mason~\cite{mason86}. Mason studies the mechanics of pushing a planar object on a horizontal plane with a point pusher. He proposes the \emph{motion cone} as a set of pushes that lead to sticking contact between the pusher and the object. Our recent work~\cite{ChavanDafle2018b} extends the idea of motion cones to prehensile tasks in a more general setting and demonstrates their application for fast in-hand manipulation planning. The application of robust motion cones for regrasp planning brings robustness to in-hand manipulation planning.

\section{Problem Formulation}
\label{sec:formulation}

We present a robust in-hand manipulation technique using a simple parallel-jaw gripper. An object in a gripper is forced to a desired pose in the grasp using prehensile pushes.
We implement the prehensile pushes by pushing the object against features in the environment. If provided with a dual-arm robot or a multi-finger gripper, the second arm or the extra fingers of the gripper can be used for the pushes.

We propose a manipulation planning framework that is compatible with several of the practical implementations for the prehensile pushes. We consider that an object is grasped in a gripper that is fixed in the world. A moving pusher pushes the object to a new pose in the grasp.
In our implementation, the pusher motion is a reflection of the robot motion against the fixed environment. Now, planning the robot motion to regrasp an object is equivalent to finding a pushing strategy that forces the object to the desired pose.

We assume the following knowledge about the physical properties of the manipulation system:
\begin{itemize}
    \item[$\Cdot$] Object geometry.
    \item[$\Cdot$] Initial and goal pose of an object in a grasp, specified by the locations and geometries of each fingers contacts.
    \item[$\Cdot$] Discrete set of pusher contacts, specified by their locations and geometries.
    \item[$\Cdot$] Coefficients of friction at the pusher contacts (an approximate value is sufficient).
\end{itemize}

Our proposed planning framework works at two levels. 
At the high-level, a T-RRT$^{*}$ based planning architecture samples the configuration space of different grasps, similar to our work in \cite{ChavanDafle2018a}.
At the low level, the planning tree is grown in the direction of the sampled poses using the knowledge of local reachable poses in the form of robust motion cones. 
Exploiting the efficient dynamics propagation via motions cones, the planner rapidly explores the configuration space and generates feasible pushing strategies to move the object in the grasp.

\section{Mechanics of Fixtureless Fixturing}
\label{sec:dynamics}

In this section, we start with a brief review on fundamental contact modelling techniques that allow us to model the force-motion interaction during prehensile pushing.
Based on the contact modelling in \secref{sec:contact} and dynamics of stable prehensile pushing in \secref{sec:pushing_dyn}, we discuss the mechanics of fixtureless fixturing and characterize the set of fixtureless fixturing pushes in \secref{sec:fixture_dyn} and \ref{sec:robust_motioncone}. 

\subsection{Contact Modelling}
\label{sec:contact}

In this paper, we assume that all bodies are rigid and all contacts have dry isotropic friction. 

\subsubsection{Limit Surface}
\label{sec:limit_surface}
Goyal~\cite{GoyalPhD89} presented the \emph{limit surface} as the boundary of the friction wrenches (force-torque pairs) that a contact with finite area can provide. 
Howe and Cutkosky~\cite{howe96} and Xydas and Kao~\cite{xydas99} showed that the limit surface geometry can be approximated as an ellipsoid. We consider circular patch contacts at the fingers. To model the force-motion interaction at the finger contacts, we assume the ellipsoidal approximation to the limit surface, which has been shown to be computationally efficient for pushing motions~\cite{lynch96,Dogar2011,lynch15,Jiaji17b}

Let $\boldsymbol{w}_c=[f_\textnormal{x}, f_\textnormal{z}, m_\textnormal{y}]$ be a frictional wrench at a finger contact in the contact frame.
The ellipsoidal limit surface constraint can be written as: $\boldsymbol{w}_c^T A \boldsymbol{w}_c=1$, where $A=Diag(a_1^{-2}, a_2^{-2}, a_3^{-2})$.
For isotropic friction, the maximum friction force is $a_1= a_2=\mu_{c} N$, where $\mu_{c}$ is the friction coefficient, and $N$ is the the normal force at the contact. The maximum friction torque about the contact normal is $a_3=rc\mu_{c} N$, where $r$ is the radius of the finger contact and $c \in [0,1]$ is the constant from numerical integration. For a uniform pressure distribution, $c$ is about 0.6~\cite{xydas99,lynch15}.
When the object slides on the contact, the friction wrench lies on the limit surface such that the normal to the limit surface is in the direction of the object motion at the contact.
If the object twist (linear and angular velocity pair), $\boldsymbol{v_\textnormal{obj}}=[v_x, v_z, \omega_y]^T$, is known, we can find the friction wrench at the contact as:
\begin{equation}
\label{eq:vel2wrench}
 \boldsymbol{w}_c= \frac{A^{-1}\boldsymbol{v_\textnormal{obj\_c}}}{\sqrt{\boldsymbol{v_\textnormal{obj\_c}}^T A^{-1} \boldsymbol{v_\textnormal{obj\_c}}}}  = \mu_c N \boldsymbol{\overline{w}}_c
\end{equation}
where $\boldsymbol{v}_{obj\_c}$ is the twist of the object in the contact frame. Here, $\boldsymbol{v}_{obj\_c}=[v_{x\_c}, v_{z\_c}, \omega_{y\_c}]^T=\boldsymbol{{J}_\textnormal{c}} \cdot \boldsymbol{v}_{obj}$, where $\boldsymbol{{J}_\textnormal{c}}$ is the Jacobian that maps the velocity from the object frame to the contact frame.
$\boldsymbol{\overline{w}}_c=[\overline{f}_\textnormal{x}, \overline{f}_\textnormal{z}, \overline{m}_\textnormal{y}]^T$ is a unit wrench on the limit surface and is scaled by maximum linear friction at the contact ($\mu_c N$) to estimate the net frictional wrench. 
%

For the ellipsoidal limit surface assumption, the linear velocity $[v_\textnormal{x\_c}, v_\textnormal{z\_c}]^T$ of the object in the contact frame is parallel and opposite to the linear frictional force $[\overline{f}_x, \overline{f}_z]^T$ applied by the contact in the contact frame~\cite{lynch92}. Moreover, the relationship between the friction wrench and the normal to the limit surface, which defines the motion direction, sets the  following constraint  between  the  angular  velocity at the contact and the linear velocity:
\begin{equation*}
 \frac{v_\textnormal{x\_c}}{{\omega}_\textnormal{y\_c}}={(rc)^2} \frac{\overline{f}_\textnormal{x}}{\overline{m}_\textnormal{y}} \ \ and \ \
 \frac{v_\textnormal{z\_c}}{{\omega}_\textnormal{y\_c}}={(rc)^2} \frac{\overline{f}_\textnormal{z}}{\overline{m}_\textnormal{y}}
\end{equation*}

Given the friction wrench on the object from the contact ($\boldsymbol{\overline{w}}_c$), we can find the object velocity as:
\begin{equation}
\label{eq:wrench2vel}
 \boldsymbol{v}_{obj} = k\boldsymbol{\Tilde{J}_\textnormal{c}} \boldsymbol{B} \cdot \boldsymbol{\overline{w}}_c \ , \  \boldsymbol{B}= Diag (1, 1, (rc)^{-2} ), \ \ k \in {\rm I\!R}^+
\end{equation}
where $\boldsymbol{\Tilde{J}_\textnormal{c}}$ maps the object velocity from the contact frame to the object frame.

\subsubsection{Generalized Friction Cone}
\label{sec:generalized_cone}
%
\begin{figure}
\centering
 \includegraphics[scale=1.04]{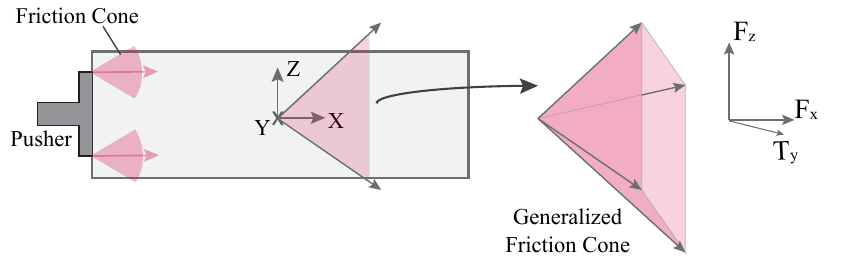}
\caption{The generalized friction cone for a line pusher modeled with point contacts at the ends. The convex hull of the set of wrenches imposed by the forces in the friction cones at the two points contacts is a polyhedral cone.}
    \label{fig:fric_cone}
\vspace{-3mm}
\end{figure} 
%
Erdmann~\cite{Erdmann94} introduced the concept of the \textit{generalized friction cone} ($\boldsymbol{W}$) as a representation of the local Coulomb friction cone at a contact in the wrench-space and in the object frame. The generalized friction cone for a patch contact modelled with multiple point contacts is the convex hull of the generalized friction cones for each constituent contact~\cite{Erdmann93}.
We model the friction at a pusher contact with its generalized friction cone: 
\begin{equation}
\label{eq:wrench_cone}
 \boldsymbol{W_\textnormal{pusher}} = \{ \boldsymbol{\overline{w}_\textnormal{pusher}} = \boldsymbol{J^\top_\textnormal{p}}\cdot \boldsymbol{\overline{f}_\textnormal{p}} \ | \ \boldsymbol{\overline{f}_\textnormal{p}} \in FC_\textnormal{pusher}\}
\end{equation}
Here, $\boldsymbol{{f}_\textnormal{p}}$ is the collection of forces at the constituent point contact/s at the pusher, and $FC_\textnormal{pusher}$ is the Coulomb friction cone/s at those constituent contact/s. The Jacobian $\boldsymbol{J^\top_\textnormal{p}}$ maps the local contact forces at the pusher ($\boldsymbol{{f}_\textnormal{p}}$) to the wrenches in the object frame; $\boldsymbol{\overline{w}_\textnormal{pusher}}$ is the unit wrench corresponding to unit force/s $\boldsymbol{\overline{f}_\textnormal{p}}$ inside the friction cone/s at the constituent contact/s of the pusher.

\figref{fig:fric_cone} shows the generalized friction cone for a line pusher modeled with point contacts at the ends. Note that the polyhedral cone form of the generalized friction cone is not an approximation. The convex hull of the wrenches exerted by the forces in the friction cones at the two points contacts in fact takes the form of a polyhedral cone.

\subsection{Dynamics of Stable Prehensile Pushing}
\label{sec:pushing_dyn}

An object in a grasp moves following the net wrench acting on the object. Under the quasi-static assumption, which is valid for slow pushing operations, the inertial forces on the object are negligible and there is a force balance:
\begin{equation}
\label{eq:force_balance}
 \boldsymbol{w_\textnormal{grasp}} +  \boldsymbol{w_\textnormal{pusher}} + m \boldsymbol{g}=0
\end{equation}

Equation~\eref{eq:force_balance} is written in the object frame located at the center of gravity of the object, where $\boldsymbol{w_\textnormal{grasp}}$ is the friction wrench provided by the grasp; $\boldsymbol{w_\textnormal{pusher}}$ is the wrench exerted by the pusher; $m$ is the mass of the object; and $\boldsymbol{g}$ is the gravitational wrench. 

This paper focuses on manipulations in a parallel-jaw grasp, so the object motions are restricted in the plane of the grasp.
We assume uniform pressure distribution at the finger contacts. 
The normal forces at the finger contacts balance each other. All forces involved in these manipulations (friction forces at the fingers, the pusher forces, and the gravitational force) are in the plane of the grasp. 
Both the fingers contribute equally to constitute the grasp wrench, i.e., $\boldsymbol{w_\textnormal{grasp}}=2 (\boldsymbol{J^\top_\textnormal{c}} \cdot \boldsymbol{{w}_\textnormal{c}})$. Here, $\boldsymbol{J^\top_\textnormal{c}}$ maps the contact wrench from the finger contact frame to the object frame.  We can rewrite \eref{eq:force_balance} as:
\begin{equation}
\label{eq:force_balance_prepush}
 2(\mu_\textnormal{c} N \boldsymbol{J^\top_\textnormal{c}} \cdot \boldsymbol{\overline{w}_\textnormal{c}}) +  \boldsymbol{J^\top_\textnormal{p}}\cdot \boldsymbol{f_\textnormal{p}} + m\boldsymbol{g} = 0
\end{equation}

A prehensile push for which the pusher contact sticks to the object during the push is called a stable prehensile push~\cite{ChavanDafle2018a}. For a stable prehensile push, constraint \eref{eq:force_balance_prepush} is balanced by the force/s inside the friction cone/s at the pusher constituent contact/s:
\begin{equation*}
\begin{split}
  2(\mu_\textnormal{c} N \boldsymbol{J^\top_\textnormal{c}} \cdot \boldsymbol{\overline{w}_\textnormal{c}}) +  \boldsymbol{J^\top_\textnormal{p}}\cdot \boldsymbol{f_\textnormal{p}} + m\boldsymbol{g} = 0 \ , \ \boldsymbol{{f}_\textnormal{p}} \in FC_\textnormal{pusher}
\end{split}
\end{equation*}
%
%
Using \eref{eq:wrench_cone}, we can rewrite the previous equation as:
\begin{equation}
\label{eq:stable_check_prepush1}
\begin{split}
  -2(\mu_\textnormal{c} N \boldsymbol{J^\top_\textnormal{c}} \cdot \boldsymbol{\overline{w}_\textnormal{c}}) - m\boldsymbol{g} = k_1 \boldsymbol{\overline{w}_\textnormal{pusher}} \\ 
  \boldsymbol{\overline{w}_\textnormal{pusher}} \in \boldsymbol{W_\textnormal{pusher}} \ , \ k_1 \in {\rm I\!R}^+
\end{split}
\end{equation}
where $k_1$ is the magnitude of the pusher normal force. 
To find if a particular object motion can be achieved with a stable push, we can check if the net required wrench falls inside the generalized friction cone of the pusher:
\begin{equation}
\label{eq:stable_check_prepush2}
\begin{split}
  -2(\mu_\textnormal{c} N \boldsymbol{J^\top_\textnormal{c}} \cdot \boldsymbol{\overline{w}_\textnormal{c}}) - m\boldsymbol{g} \in \boldsymbol{W_\textnormal{pusher}}
\end{split}
\end{equation}
%


\subsection{Dynamics of Fixtureless Fixturing}
\label{sec:fixture_dyn}
We define \emph{fixtureless fixturing} as a subset of stable prehensile pushes that are invariant to the object inertia, grasping force and friction parameters at the finger contacts.

The presence of gravity in \eref{eq:stable_check_prepush1} makes the dynamics of prehensile pushing sensitive not only to the object inertia ($m\boldsymbol{g}$), but also to the  grasping force ($N$) and the coefficient of friction at the fingers ($\mu_\textnormal{c}$). 
%
If the pusher is aligned such that the pusher contact normal is along the direction of gravity, the gravitational force on the  object is entirely balanced by the part of the normal force at the pusher. 
Then the dynamics check for stable prehensile pushing \eref{eq:stable_check_prepush1} becomes:
\begin{equation*}
\begin{split}
  -2(\mu_\textnormal{c} N \boldsymbol{J^\top_\textnormal{c}} \cdot \boldsymbol{\overline{w}_\textnormal{c}}) - k_2\boldsymbol{\overline{w}_\textnormal{pusher\_n}} = k_1 \boldsymbol{\overline{w}_\textnormal{pusher}} \\ 
  \boldsymbol{\overline{w}_\textnormal{pusher}} \in \boldsymbol{W_\textnormal{pusher}} \ , \ \ \ k_1, k_2 \in {\rm I\!R}^+
\end{split}
\end{equation*}
\begin{equation*}
\begin{split}
\vspace{-1mm}
  -2(\mu_\textnormal{c} N \boldsymbol{J^\top_\textnormal{c}} \cdot \boldsymbol{\overline{w}_\textnormal{c}}) = k_1 \boldsymbol{\overline{w}_\textnormal{pusher}} + k_2\boldsymbol{\overline{w}_\textnormal{pusher\_n}} \\ 
  \boldsymbol{\overline{w}_\textnormal{pusher}} \in \boldsymbol{W_\textnormal{pusher}} \ , \ \ \ k_1, k_2 \in {\rm I\!R}^+
\end{split}
\end{equation*}
%
%
where $\boldsymbol{\overline{w}_\textnormal{pusher\_n}}$ is the unit normal wrench at the pusher contact. We know that $\boldsymbol{\overline{w}_\textnormal{pusher\_n}} \in \boldsymbol{W_\textnormal{pusher}}$, so we can rewrite the above constraint as:
\begin{equation*}
\vspace{-1mm}
\begin{split}
  -2(\mu_\textnormal{c} N \boldsymbol{J^\top_\textnormal{c}} \cdot \boldsymbol{\overline{w}_\textnormal{c}}) = k_3 \boldsymbol{\overline{w}_\textnormal{p}} \\ 
  \boldsymbol{\overline{w}_\textnormal{p}} \in \boldsymbol{W_\textnormal{pusher}} \ , \ \ \ k_3 \in {\rm I\!R}^+
\end{split}
\end{equation*}
\noindent As $\mu_\textnormal{c}$ and $N$ are scalar constants, we further simplify it as:
\begin{equation}
\label{eq:stable_check_prepush_gfree1}
\begin{split}
  -\boldsymbol{J^\top_\textnormal{c}} \cdot \boldsymbol{\overline{w}_\textnormal{c}} = k_4 \boldsymbol{\overline{w}_\textnormal{p}} \\ 
  \boldsymbol{\overline{w}_\textnormal{p}} \in \boldsymbol{W_\textnormal{pusher}} \ , \ \ \ k_4 \in {\rm I\!R}^+
\end{split}
\end{equation}
Now, we can write the dynamics condition for stable prehensile pushing with gravity balancing pushers as: 
\begin{equation}
\label{eq:stable_check_prepush_gfree2}
\begin{split}
  -\boldsymbol{J^\top_\textnormal{c}} \cdot \boldsymbol{\overline{w}_\textnormal{c}} \in \boldsymbol{W_\textnormal{pusher}}
\end{split}
\end{equation}
\begin{figure}
\centering
 \includegraphics[scale=1]{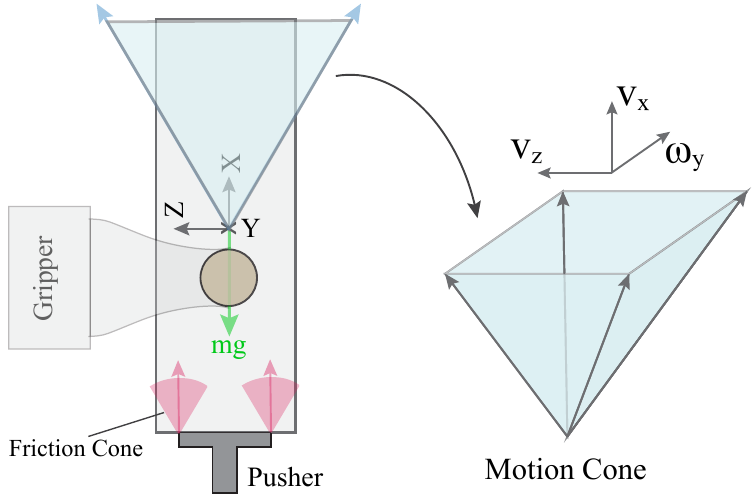}
\caption{An example of a robust motion cone. Note that the motion cone is a three-dimensional cone in the space of the linear velocity of the object in the $XZ$ plane and the rotational velocity of the object about $Y$ axis.}
    \label{fig:motion_cone}
\vspace{-3mm}
\end{figure} 

Note the invariance of \eref{eq:stable_check_prepush_gfree1} and \eref{eq:stable_check_prepush_gfree2} on the object inertia, grasping force and the friction coefficient at the fingers.
All the object motions that satisfy \eref{eq:stable_check_prepush_gfree2} can be achieved with fixtureless fixturing when pushing in the gravity-balancing orientation. For planning regrasps with fixtureless fixturing, rather than checking different instantaneous object motions for their validity for \eref{eq:stable_check_prepush_gfree2}, a direct bound on the feasible object motions is more efficient. 

\subsection{Computation of Robust Motion Cone}
\label{sec:robust_motioncone}

A motion cone abstracts the dynamics of pushing and provides direct bounds on the object motions that can be achieved while keeping the pusher contact sticking to the object~\cite{mason86}.
In this section, we will extend the idea of the motion cone for fixtureless fixturing.

\textbf{\textit{Problem}}: Find a set of instantaneous object motions that can be achieved with fixtureless fixturing.

This is equivalent to finding a set of object motions that satisfy constraint \eref{eq:stable_check_prepush_gfree2}. 
To find a motion cone, we find a set of support contact wrenches ($\boldsymbol{\overline{w}_\textnormal{c}}$) that satisfy \eref{eq:stable_check_prepush_gfree2} and then map this wrench-set ($\boldsymbol{\overline{W}_\textnormal{c}}$) to a set of object twists. We denote this object twist-set by $\boldsymbol{\overline{V}_\textnormal{obj}}$ and call it \emph{robust motion cone}.

From the constraint \eref{eq:stable_check_prepush_gfree2} we can observe that $-\boldsymbol{J^\top_\textnormal{c}} \cdot \boldsymbol{\overline{W}_\textnormal{c}}=\boldsymbol{W_\textnormal{pusher}}$ or  $\boldsymbol{\overline{W}_\textnormal{c}}=-\boldsymbol{\Tilde{J}^\top_\textnormal{c}} \cdot \boldsymbol{W_\textnormal{pusher}}$ where $\boldsymbol{\Tilde{J}^\top_\textnormal{c}}$ maps wrenches from the object frame to the finger contact frame. From definition \eref{eq:wrench_cone}, $\boldsymbol{W_\textnormal{pusher}}$ is convex and polyhedral. Therefore, $\boldsymbol{\overline{W}_\textnormal{c}}$ and $\boldsymbol{\overline{V}_\textnormal{obj}}$ are also convex and polyhedral.
The generators of the robust motion cone $\boldsymbol{\overline{V}_\textnormal{obj}}$ can be computed by a linear mapping of the generators of the generalized friction cone of the pusher $\boldsymbol{W_\textnormal{pusher}}$. \figref{fig:motion_cone} shows a graphical illustration of the motion cone computed for a pushing scenario.

\myparagraph{Procedure to compute a robust motion cone}:
\begin{enumerate}
    \item Solve $\boldsymbol{\overline{w}_\textnormal{c}}=-\boldsymbol{\Tilde{J}^\top_\textnormal{c}} \cdot \boldsymbol{\overline{w}_\textnormal{p}}$ for  $\boldsymbol{\overline{w}_\textnormal{p}}$ equal to every edge of the generalized friction cone $\boldsymbol{W_\textnormal{pusher}}$.
    
    \item Define the set of  $\boldsymbol{\overline{w}_\textnormal{c}}$ computed in step 1 as the generators/edges of the wrench-set $\boldsymbol{\overline{W}_\textnormal{c}}$.
    
    \item Map $\boldsymbol{\overline{W}_\textnormal{c}}$ to the object twist space using \eref{eq:wrench2vel} to obtain the robust motion cone $\boldsymbol{\overline{V}_\textnormal{obj}}$.
\end{enumerate}

\section{Application to Regrasp Planning}
\label{sec:planning}

In this section, we discuss the application of fixtureless fixturing to regrasp planning. We present a sampling-based planning framework that uses the knowledge of robust motion cones and generates a tree of grasps that can be reached with fixtureless fixturing. A path in this tree is a pushing strategy to regrasp an object in the gripper through a series of robust stable prehensile pushes.

The high-level planning framework is similar to that presented in our prior work~\cite{ChavanDafle2017,ChavanDafle2018a, ChavanDafle2018b}. It is based on T-RRT$^*$ -- an optimal sampling based method for planning on configuration space cost-maps~\citep{trrt_star,trrt}.

\myparagraph{For selective sampling}, the T-RRT* framework uses a transition test and filters the sampled configurations to prefer the exploration in low configuration-cost regions. We define the configuration cost as the distance from the goal. The transition test\footnote{The implementation of the transition test and its adaptive nature is discussed in detail in~\cite{ChavanDafle2017}} gives preference to the stochastic exploration towards the goal grasp, while allowing the flexibility to move the object away from the goal if that is necessary to move the object finally to the goal.

\myparagraph{For effective connections}, the T-RRT$^*$ exploits the underlying RRT$^*$~\citep{rrt_star} framework to make and rewire the connections in the tree such that the cost of the nodes is reduced.
We define the cost of a node as the sum of the cost of the parent node and the cost to push the object to the sampled node. The cost of a push is $0.1$ if the parent node uses the same pusher as the child and $1$ otherwise.  

In practice, every time a pusher contact is changed, it introduces a possibility to add noise to the system. Reducing the pusher switch-overs can add to the robustness of the manipulations planned. 
With our node cost definition, the planner tries to reduce the number of  pusher switch-overs required to push the object to the desired pose.

\begin{algorithm}[t]
  \caption{: In-Hand Manipulation Planner}\label{alg:full_planner}
  $  \textbf{input}: q_{init}, q_{goal}$ \par
  $  \textbf{output}:$ {tree} $\ \mathcal{T}$
  \begin{algorithmic}
  \State $\mathcal{T}\gets \textrm{initialize tree}(q_{init})$
  \State $ \textrm{generate\_robustCones}(\mathcal{T},q_{init})$
      \While{$q_{goal} \notin \mathcal{T}$ \textbf{or} cost($q_{goal}) > \textrm{cost threshold} $}
        \State $q_{rand}\gets \textrm{sample random configuration}(\mathcal{C})$
        
        \State $q_{parent}\gets \textrm{find nearest neighbor}(\mathcal{T},q_{rand})$
        \State $q_{sample}\gets \textrm{take unit step}(q_{parent},q_{rand})$
        \If{$q_{sample} \notin \mathcal{T}$}
        
        \If{\textrm{transition test}$(q_{parent},q_{sample},\mathcal{T})$}
            \State $q_{new} \gets \textrm{robust\_push}(q_{parent},q_{sample})$
            
            \If{\textrm{transition test}$(q_{parent},q_{new},\mathcal{T})$ \textbf{and} \\ \hspace{19mm}\textrm{grasp maintained}$(q_{new})$} 

                \State $q\mbox{*}_{parent}\gets \textrm{optimEdge}(\mathcal{T},q_{new},q_{parent})$
                
                \State $\textrm{add new node}(\mathcal{T},q_{new})$
                
                \State $\textrm{add new edge}(q\mbox{*}_{parent},q_{new})$
                
                \State $ \textrm{generate\_robustCones}(\mathcal{T},q_{new})$
                
                \State $\textrm{rewire tree}(\mathcal{T},q_{new},q\mbox{*}_{parent})$
            \EndIf
            \EndIf
        \EndIf
    \EndWhile
  \end{algorithmic}
\end{algorithm}
Algorithm \ref{alg:full_planner} shows our in-hand manipulation planner.
Let $q$ be a configuration of an object, i.e., the pose of the object in the gripper frame that is fixed in the world.
In this paper, we consider manipulations in a parallel-jaw grasp, so the configuration space $\mathcal{C}$ is $[X, Z, \theta_y] \in {\rm I\!R}^3$. The object can translate in the grasp plane ($XZ$ plane) and rotate around a perpendicular to the grasp plane ($Y$ axis).

Let $q_{init}$ and $q_{goal}$ be an initial and desired pose of the object in the grasp frame, respectively. 
The planner initiates a tree $\mathcal{T}$ with $q_{init}$ and generates robust motion cones at it. 
While the goal pose is not reached within some cost threshold, a random configuration ($q_{rand}$) is sampled. A nearest configuration ($q_{parent}$) to $q_{rand}$ in the tree $\mathcal{T}$ is found, and an unit-step object pose ($q_{sample}$) towards the $q_{rand}$ is computed. 
Using the transition test, the planner evaluates whether a move from the parent node $q_{parent}$ to $q_{sample}$ is beneficial or not.
If it is beneficial, the \textit{robust\_push} routine computes an object configuration ($q_{new}$) closest to $q_{sample}$ that can be reached using the robust motion cones at $q_{parent}$. The node $q_{new}$ is added to the tree such that the node costs of $q_{new}$ and the nodes near $q_{new}$ are lowered if possible. The robust motion cones are generated for every new node added to the tree.

\textit{generate\_robustCones} and \textit{robust\_push} are two important routines for the regrasp planning using fixtureless fixturing:

\myparagraph{\textit{generate\_robustCones}} computes polyhedral robust motion cones for a given object configuration in the grasp using the procedure explained in \secref{sec:robust_motioncone}. At every node, the number of motion cones is the same as that of the pushers.

\myparagraph{\textit{robust\_push}} finds an object pose closest to the sampled pose ($q_{sample}$) that can be reached with fixtureless fixturing. This computation is done using the robust motion cones at the parent node ($q_{parent}$), which provide the knowledge of local reachability.
If the object twist needed from $q_{parent}$ to $q_{sample}$ is inside any of the robust motion cones, the sampled pose can be reached from $q_{parent}$.
If the required object twist is outside all the motion cones, a twist that is inside one of the motion cones and closest to the desired twist is selected.

The motion cone based unit-step propagation and T-RRT$^*$-based high-level framework work together to rapidly explore the configuration space of different grasps and generate pushing strategies for the desired in-hand manipulation. The manipulation plans generated by the planner are invariant to the object inertia, grasping force and friction at the fingers.

\section{Regrasp Examples and Experimental Validation}
\label{sec:Exp_examples}
\begin{figure}
\centering
 \includegraphics[scale=0.7]{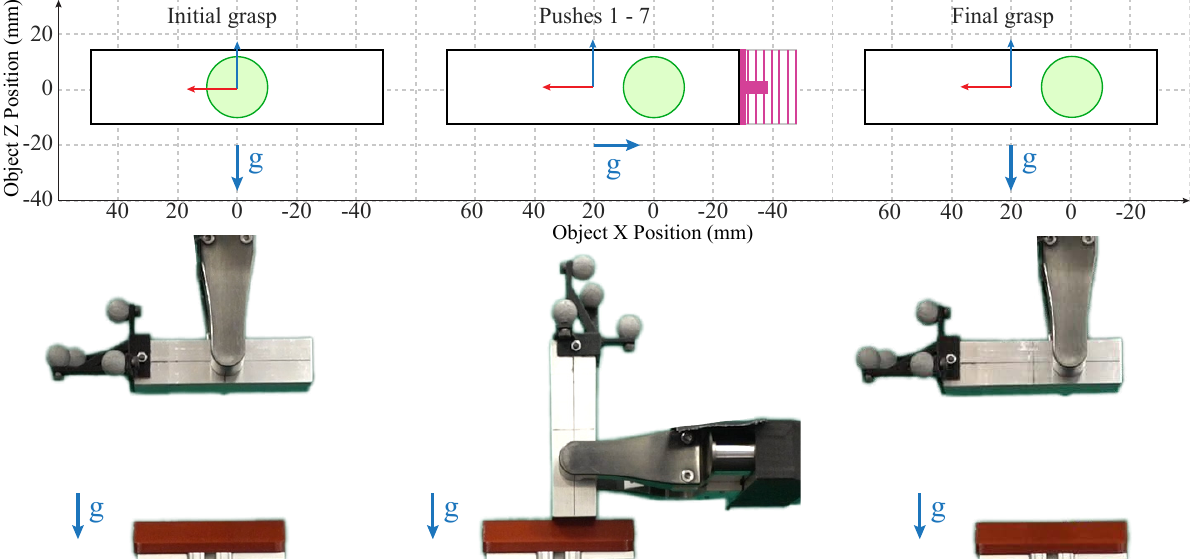}
\caption{Simulated motion of the object and snapshots of the experimental run  for  a  pushing strategy to offset the object in the grasp using low coefficient pushers. The circular finger contact is shown in green, while the line pusher contact is shown in magenta.}
    \label{fig:linpush}
\end{figure} 
\begin{table}[b]
\vspace{-3mm}
  \caption{Physical properties of the objects used.}
  \label{tab:objects}
	\centering
	\begin{tabular}{|l|l|c|r|}
         \hline
          \textbf{Shape} & \textbf{Material} & \textbf{Dim. [L, B, H]} (mm) & \textbf{Mass} (g) \\ \hline
          square prism  & Al 6061 & 100, 25, 25 & 202\\ \hline  
          rectangular prism  & Delrin & 80, 25, 38 & 130\\ \hline
          T-shaped & ABS  & 70, 25, 50 & 62\\ \hline
    \end{tabular}
\end{table}
We validate our manipulation planner with regrasping tasks in simulation and with experiments. The experiments were performed on a manipulation platform instrumented with an industrial robot arm, a parallel-jaw gripper with force control, and a Vicon system for object tracking.

\subsection{Simulation and Experimental Setup}
\label{sec:setup}
We consider different regrasp tasks for the objects listed in Table \ref{tab:objects}. We use a parallel-jaw gripper with flat circular finger contacts. The planner is initiated with pusher contacts on either sides of the object and under the object. In practice, all these contacts are implemented using a single feature in the environment. 

The initial pose of the object is treated as $[0,0,0]$, and the goal poses for different regrasp examples are listed in \tabref{tab:timing}. The plan time for each regrasp, shown in \tabref{tab:timing}, is the median time in seconds over 10 planning trials. A computer with an Intel Core i7 2.8 GHz processor and MATLAB R2017a was used for all the computations and planning.

\subsection{Regrasp Examples}
\label{sec:examples}
\subsubsection{Regrasping an object offset to the center}
The goal in this example is to regrasp the square prism 20 mm offset from the center in the horizontal direction. The gripping force, and frictional parameters at the fingers and at the features in the environment, are chosen such that pushing the object horizontally from side will not be a valid solution. Due to the gravitational force, the object will also slide downwards by a few millimeters if pushed from the side, as observed in~\cite{Kolbert16}. 
\begin{table}[b]
\vspace{-4mm}
\caption{Planning times (in seconds) with different methods for unit-step propagation}
  \label{tab:timing}
	\centering
	\begin{tabular}{|c|c|c|c|c|}
         \hline
          \textbf{Manipulation} & \textbf{Goal} & \textbf{Plan}  & \textbf{Plan}   & \textbf{Plan}\\
           & [$X, Z, \theta_Y $] & \textbf{Time} & \textbf{Time}~\cite{ChavanDafle2018a} & \textbf{Time}~\cite{ChavanDafle2017} \\
          \hline
          Horz. offset (low $\mu$)  & 20, 0, 0 & 0.39 & 2.83 & 592.8\\ \hline
          [$X, Z, \theta_Y$] Regrasp & 15, -13, 45 & 0.65 & 2.54 & 17684\\ \hline
          T-shaped & 25, 17.5, 0 & 0.60 & 0.82 & 32657\\ \hline
    \end{tabular}
\end{table}
\begin{figure}
\centering
 \includegraphics[scale=0.7]{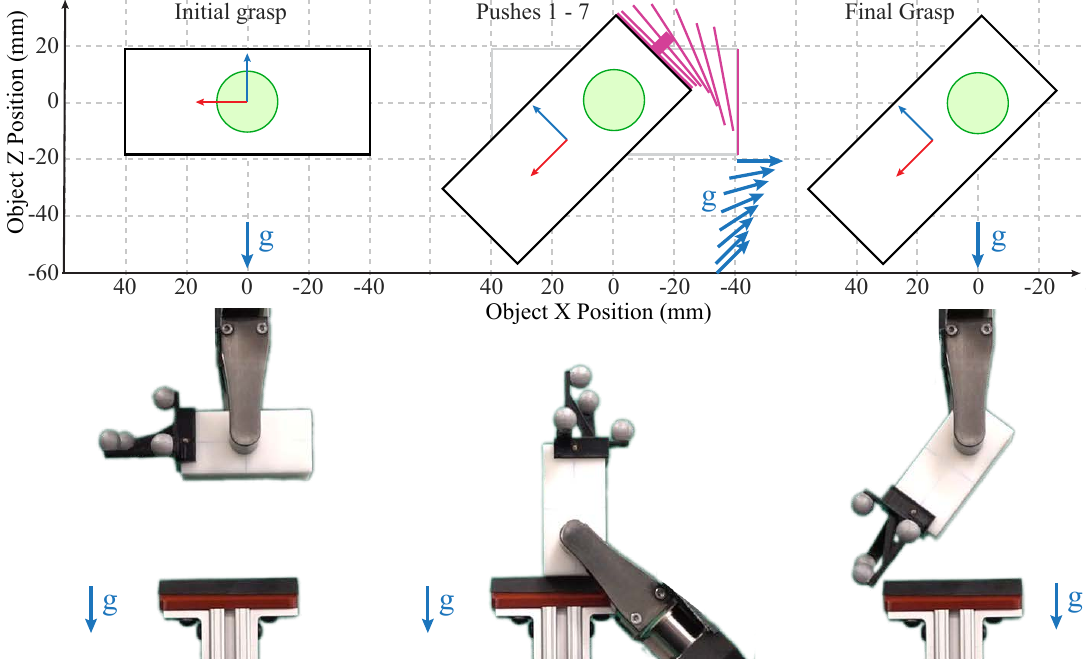}
\caption{Simulated motion of the object and snapshots of the experimental run for a general regrasp in [$X, Z, \theta_y$]. In the simulation, the direction of gravity remains constant in the pusher frame because, in reality, the pusher is a fixed feature in the environment}
    \label{fig:prepush}
    \vspace{-5mm}
\end{figure} 

The planner proposed in this paper rotates the gripper such that the object is pushed from a side face but using the feature located horizontally,  as shown in \figref{fig:linpush}. The gravitational force on the object is along the pusher contact normal and the regrasp is achieved using fixtureless fixturing. This strategy is simpler and more robust than the solutions that the planners in~\cite{ChavanDafle2017,ChavanDafle2018a} generated for the same problem. 
In~\cite{ChavanDafle2017} and~\cite{ChavanDafle2018a}, the planner uses multiple pushers, specifically one from side and one from bottom face, to achieve this regrasp.
With the planner proposed in this paper, no contact switch-over is needed for the desired regrap as shown in \figref{fig:linpush}.
Moreover, the time taken to plan such a strategy is 5 to 1500 times faster than our prior work as shown in Table \ref{tab:timing}.

\subsubsection{General manipulation in $[X, Z, \theta_y]$}

This example demonstrates large in-hand manipulation for the rectangular prism, involving both displacement and rotation of the fingers on the object. The planner quickly converges to a strategy which does not require any pusher switch-over. The knowledge of motion cones allows the planner to reason about all feasible motions that can be made instantaneously and to choose the motion that best moves the object towards the goal. \figref{fig:prepush} shows the pushing strategy our planner generated.

\subsubsection{Manipulation with a complex non-convex object}
\begin{figure}
\centering
 \includegraphics[scale=0.85]{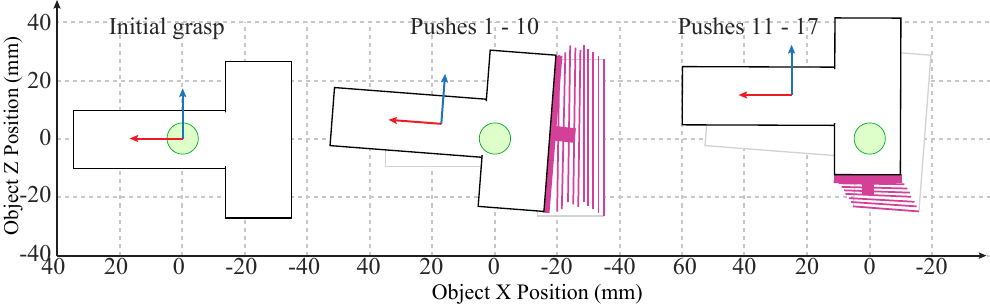}
\caption{Simulated motion of the object for a pushing strategy for manipulating a T-shaped object. \figref{fig:tpush_exp} shows the experimental run.}
    \label{fig:tpush_sim}
    \vspace{-3mm}
\end{figure}

Here we show manipulation of a T-shaped object. Manipulating such a non-convex object is particularly hard, because pushing the object directly towards the goal may lead to losing the grasp on the object. The long horizon planning feature of the planner and the capability to reason about discreetly changing the pushers are essential for regrasping this object.
As shown in \figref{fig:tpush_exp} and \figref{fig:tpush_sim}, our planner generates a strategy that respects the object shape and corresponding geometric constraints. The planner utilizes pushes from two different sides to finally move the object to the goal grasp. 
%

\subsection{Experimental Observations}
\label{sec:observations}
In this section, we highlight some of the key characteristics of fixtureless fixturing that are supported experimentally. We carried out multiple runs of the pushing strategies in \secref{sec:examples} for different grasping forces and recorded the object motion in the grasp. 

\subsubsection{Minimal displacement of the object}

The experimental data shows that the object moves a minimal amount with respect to the environment during the planned regrasps. From the mechanics of robust stable prehensile pushing, we expect the object to stick to the environment during the regrasps. 

\figref{fig:linpush_error} shows the displacement of the object observed during the pushing strategy shown in \figref{fig:linpush}. 
The displacement in the X direction is less than $\pm 0.2$~mm, whereas that in the Z direction is even smaller, less than $\pm 0.1$~mm.
These values are close to the precision level of the Vicon object tracking system, so some of the object displacement can be attributed to the noise from the camera system.

\figref{fig:prepush_error} shows the displacement of the object during the pushing strategy shown in \figref{fig:prepush}. 
The object moves during the regrasp by a very small amount, less than $3\%$ of the object displacement required for this regrasp.

\subsubsection{Robustness against variation in the friction at the fingers}
\label{sec:robustness}
\begin{figure}[t]
\centering
 \includegraphics[scale=0.98]{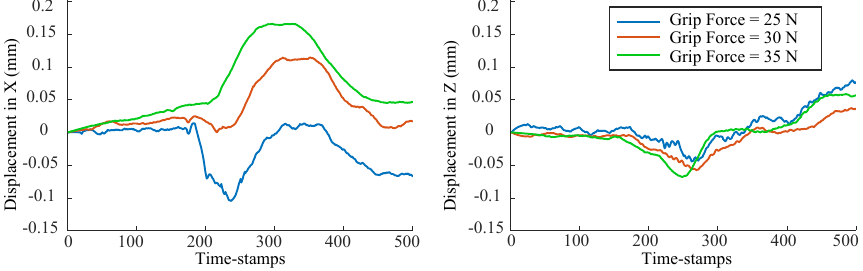}
\caption{Displacement of the object with respect to the environment for the regrasp action shown in \figref{fig:linpush}. As expected, during fixtureless fixturing, the object sticks to the environment and moves by a negligible amount as the fingers slide on it.}
    \label{fig:linpush_error}
\end{figure}

\begin{figure}
\centering
 \includegraphics[scale=0.98]{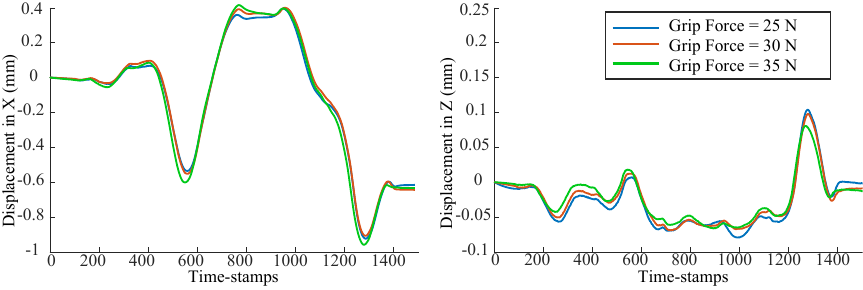}
\caption{Displacement of the object with respect to the environment for the regrasp action shown in \figref{fig:prepush}. 
}
\label{fig:prepush_error}
\vspace{-3mm}
\end{figure}

We observe a minimal effect on the outcome of the regrasp actions planned with fixtureless fixturing when the friction at the fingers is changed.
As the grasping force changes, the friction force at the fingers change. However, as seen in the plots in \figref{fig:linpush_error} and \figref{fig:prepush_error}, the variation in the outcome of the pushing strategy for three different grasping forces is negligible. This supports the invariance we expect theoretically from fixtureless fixturing.

The repeatability and robustness of the pushing actions allowed the robot to place the T-shaped object after every regrasp and restart the next run autonomously twenty times in a row. The T-shaped object does not have Vicon markers on it for tracking. Twenty runs for the manipulation shown in \figref{fig:tpush_exp} were performed in a sequence in an open-loop fashion with different grasping forces. 

\subsubsection{Universal Fixture} For all the examples discussed in \secref{sec:examples}, a single feature in the environment was sufficient to manipulate different objects. Fixtureless fixturing allows robots to use a simple feature as a ``universal'' fixture for different objects and different regrasping tasks.

\section{Discussion}
\label{sec:discussion}
We present a manipulation method that acts as a fixturing technique for regrasping without using a conventional fixturing hardware. The object in a gripper is manipulated with a series of pushes against simple features in the environment such that the environment keeps the object stationary while the fingers slide on the object to a desired pose. We refer to this approach as \emph{fixtureless fixturing}. 

Based the mechanics of prehensile pushing, we construct convex polyhedral sets of object motions that can be produced with fixtureless fixturing. We call this subsets of pushes robust motion cones and show that they are invariant to the object inertia, gripping force, and friction at the fingers. The robust motion cones abstract the dynamics of prehensile pushing and provide direct bounds on the object motions that are feasible with fixtureless fixturing.

We demonstrate the application of the robust motion cones for planning in-hand manipulations with a parallel-jaw gripper.  Our planner generates pushing strategies in a fraction of a second to force the object to the goal pose. The generated pushing plans are robust against the uncertainty in the object inertia, friction at the fingers, and gripping force.
We validate this robustness with controlled experiments on a platform instrumented with a Vicon object tracking system. 
The experimental results confirm that the object remains fixtured to the environment during the manipulations. Moreover, minimal variation is observed in the outcome of the regrasp experiments performed with different grasping forces.

Using the fixtureless fixturing technique, robots can exploit simple features in the environment as versatile fixtures. The flexibility gained by using fixtureless fixturing facilitates dexterous manipulation even with simple grippers. With such an approach to in-hand manipulation, we move a step closer to the goal of versatile, yet simple, robust, and flexible robotic automation.


\bibliographystyle{IEEEtranN} 
{\small \bibliography{ncd-case18}}

\begin{thebibliography}{10}
\providecommand{\url}[1]{#1}
\csname url@rmstyle\endcsname
\providecommand{\newblock}{\relax}
\providecommand{\bibinfo}[2]{#2}
\providecommand\BIBentrySTDinterwordspacing{\spaceskip=0pt\relax}
\providecommand\BIBentryALTinterwordstretchfactor{4}
\providecommand\BIBentryALTinterwordspacing{\spaceskip=\fontdimen2\font plus
\BIBentryALTinterwordstretchfactor\fontdimen3\font minus
  \fontdimen4\font\relax}
\providecommand\BIBforeignlanguage[2]{{%
\expandafter\ifx\csname l@#1\endcsname\relax
\typeout{** WARNING: IEEEtran.bst: No hyphenation pattern has been}%
\typeout{** loaded for the language `#1'. Using the pattern for}%
\typeout{** the default language instead.}%
\else
\language=\csname l@#1\endcsname
\fi
#2}}

\bibitem{ChavanDafle2014}
N.~{Chavan-Dafle}, A.~Rodriguez, R.~Paolini, B.~Tang, S.~S. Srinivasa,
  M.~Erdmann, M.~T. Mason, I.~Lundberg, H.~Staab, and T.~Fuhlbrigge,
  ``\href{http://ieeexplore.ieee.org/document/6907062/}{Extrinsic dexterity:
  In-hand manipulation with external forces},'' in \emph{IEEE International
  Conference on Robotics and Automation (ICRA)}, 2014, pp. 1578--1585.

\bibitem{ChavanDafle2015a}
N.~{Chavan-Dafle} and A.~Rodriguez,
  ``\href{http://ieeexplore.ieee.org/document/7354264/?reload=true}{Prehensile
  Pushing: In-hand Manipulation with Push-Primitives},'' in \emph{IEEE/RSJ
  International Conference on Intelligent Robots and Systems (IROS)}, 2015, pp.
  6215 -- 6222.

\bibitem{ChavanDafle2017}
N.~{Chavan-Dafle} and A.~Rodriguez\vspace{0.1mm},
  ``\href{https://arxiv.org/pdf/1707.00318.pdf}{Sampling-based Planning of
  In-Hand Manipulation with External Pushes},'' in \emph{Int Symp Robot Res},
  2017.

\bibitem{ChavanDafle2018a}
N.~{Chavan-Dafle} and A.~Rodriguez\vspace{0mm},
  ``\href{https://arxiv.org/pdf/1710.11097.pdf}{Stable Prehensile Pushing:
  In-Hand Manipulation with Alternating Sticking Contacts},'' in \emph{IEEE
  International Conference on Robotics and Automation (ICRA)}, 2018.

\bibitem{american1962handbook}
{American Society of Tool and Manufacturing Engineers}, \emph{Handbook of
  fixture design: a practical reference book of workholding principles and
  designs for all classes of machining, assembly, and inspection}.\hskip 1em
  plus 0.5em minus 0.4em\relax McGraw-Hill, 1962.

\bibitem{Asada85}
H.~Asada and A.~By,
  ``\href{http://ieeexplore.ieee.org/document/1087007/?reload=true}{Kinematic
  analysis of workpart fixturing for flexible assembly with automatically
  reconfigurable fixtures},'' \emph{IEEE Journal on Robotics and Automation},
  vol.~1, no.~2, pp. 86--94, 1985.

\bibitem{Chou89}
Y.-C. Chou, V.~Chandru, and M.~Barash,
  ``\href{http://manufacturingscience.asmedigitalcollection.asme.org/article.aspx?articleid=1447185}{A
  Mathematical Approach to Automatic Configuration of Machining Fixtures:
  Analysis and Synthesis},'' \emph{ASME Journal of Engineering for Industry},
  vol. 111, no.~4, pp. 299--306, 1989.

\bibitem{Hong91}
L.~S. Hong and M.~R. Cutkosky,
  ``\href{http://manufacturingscience.asmedigitalcollection.asme.org/article.aspx?articleid=1447458}{Fixture
  Planning With Friction},'' \emph{Journal of Manufacturing Science and
  Engineering}, vol. 113(3), pp. 320--327, 1991.

\bibitem{GoyalPhD89}
S.~{Goyal},
  ``\href{http://ruina.tam.cornell.edu/research/topics/friction_and_fracture/GoyalPhDThesis.pdf}{Planar
  sliding of a rigid body with dry friction: Limit surfaces and dynamics of
  motion},'' PhD Dissertation, Dept. of Mechanical Engineering, Cornell
  University, 1989.

\bibitem{mason86}
M.~T. Mason,
  ``\href{http://journals.sagepub.com/doi/pdf/10.1177/027836498600500303}{Mechanics
  and Planning of Manipulator Pushing Operations},'' \emph{The International
  Journal of Robotics Research}, vol.~5, no.~3, pp. 53--71, 1986.

\bibitem{ChavanDafle2018b}
N.~Chavan-Dafle, R.~Holladay, and A.~Rodriguez, ``In-hand manipulation via
  motion cones,'' in \emph{Robotics: Science and Systems (RSS)}, 2018.

\bibitem{howe96}
R.~D. Howe and M.~R. Cutkosky,
  ``\href{https://doi.org/10.1177/027836499601500603}{Practical Force-Motion
  Models for Sliding Manipulation},'' \emph{The International Journal of
  Robotics Research}, vol.~15, no.~6, pp. 557--572, 1996.

\bibitem{xydas99}
N.~Xydas and I.~Kao,
  ``\href{https://doi.org/10.1177/02783649922066673}{Modeling of Contact
  Mechanics and Friction Limit Surfaces for Soft Fingers in Robotics, with
  Experimental Results},'' \emph{The International Journal of Robotics
  Research}, vol.~18, no.~9, pp. 941--950, 1999.

\bibitem{lynch96}
K.~M. Lynch and M.~T. Mason,
  ``\href{http://journals.sagepub.com/doi/pdf/10.1177/027836499601500602}{Stable
  Pushing: Mechanics, Controllability, and Planning},'' \emph{The International
  Journal of Robotics Research}, vol.~15, no.~6, pp. 533--556, 1996.

\bibitem{Dogar2011}
M.~Dogar and S.~Srinivasa,
  ``\href{http://www.ri.cmu.edu/pub_files/2011/7/DogarAndSrinivasa_RSS2011.pdf}{A
  Framework for Push-grasping in Clutter},'' in \emph{Robotics: Science and
  Systems}.\hskip 1em plus 0.5em minus 0.4em\relax MIT Press, July 2011.

\bibitem{lynch15}
J.~Shi, J.~Z. Woodruff, and K.~M. Lynch,
  ``\href{http://ieeexplore.ieee.org/abstract/document/7353474/citations}{Dynamic
  in-hand sliding manipulation},'' in \emph{2015 IEEE International Conference
  on Intelligent Robots and Systems}, 2015, pp. 870--877.

\bibitem{Jiaji17b}
J.~Zhou and M.~T. Mason,
  ``\href{https://par.nsf.gov/servlets/purl/10046035}{Pushing revisited:
  Differential flatness, trajectory planning and stabilization},'' in \emph{Int
  Symp Robot Res}, 2017.

\bibitem{lynch92}
K.~M. Lynch, H.~Maekawa, and K.~Tanie,
  ``\href{https://www.ri.cmu.edu/pub_files/pub2/lynch_kevin_1992_2/lynch_kevin_1992_2.pdf}{Manipulation
  And Active Sensing By Pushing Using Tactile Feedback},'' in \emph{Proceedings
  of the IEEE/RSJ International Conference on Intelligent Robots and Systems},
  vol.~1, 1992, pp. 416--421.

\bibitem{Erdmann94}
M.~Erdmann,
  ``\href{http://journals.sagepub.com/doi/pdf/10.1177/027836499401300306}{On a
  Representation of Friction in Configuration Space},'' \emph{International
  Journal of Robotics Research}, vol.~13, no.~3, pp. 240--271, 1994.

\bibitem{Erdmann93}
M.~Erdmann\vspace{0mm},
  ``\href{https://www.ri.cmu.edu/pub_files/pub2/erdmann_michael_1993_1/erdmann_michael_1993_1.pdf}{Multiple-point
  contact with friction: Computing forces and motions in configuration
  space},'' in \emph{IEEE Int Conf Robot and Sys}, vol.~1, 1993, pp. 163--170
  vol.1.

\bibitem{trrt_star}
D.~Devaurs, T.~Sim\'eon, and J.~Cort\'es,
  ``\href{http://ieeexplore.ieee.org/document/7305826/?reload=true&arnumber=7305826}{Optimal
  Path Planning in Complex Cost Spaces With Sampling-Based Algorithms},''
  \emph{IEEE Transactions on Automation Science and Engineering}, vol.~13,
  no.~2, pp. 415--424, 2016.

\bibitem{trrt}
L.~Jaillet, J.~Cort\'es, and T.~Sim\'eon,
  ``\href{http://ieeexplore.ieee.org/abstract/document/5477164/citations}{Sampling-Based
  Path Planning on Configuration-Space Costmaps},'' \emph{IEEE Transactions on
  Robotics}, vol.~26, no.~4, pp. 635--646, 2010.

\bibitem{rrt_star}
S.~Karaman and E.~Frazzoli,
  ``\href{http://journals.sagepub.com/doi/abs/10.1177/0278364911406761}{Sampling-based
  Algorithms for Optimal Motion Planning},'' \emph{International Journal of
  Robotics Research}, vol.~30, no.~7, pp. 846--894, 2011.

\bibitem{Kolbert16}
R.~Kolbert, N.~Chavan-Dafle, and A.~Rodriguez,
  ``\href{https://nikhilcd.mit.edu/sites/default/files/documents/2016_ISER_Pushing_Experiments-preprint.pdf}{Experimental
  Validation of Contact Dynamics for In-Hand Manipulation},'' in
  \emph{International Symposium on Experimental Robotics}, 2016.

\end{thebibliography}

\end{document}